\documentclass{article}
\usepackage{spconf,amsmath,graphicx}

\usepackage{cite}
\usepackage{amsmath,amssymb,amsfonts}
\usepackage{algorithmic}
\usepackage{graphicx}
\usepackage{textcomp}
\usepackage{xcolor}
\usepackage{flushend}

\usepackage{booktabs,makecell,tabularx}
\usepackage{multirow}
\usepackage{hyperref}
\usepackage[export]{adjustbox} %centering figure

\usepackage{xcolor}
\newcommand{\colA}[1]{\textcolor{black}{#1}} % color for Reviewer#1 -- 0ECC 
\newcommand{\colB}[1]{\textcolor{black}{#1}} % color for Reviewer#2 -- 0C7E

\newcommand{\ModelName}{Sparse R-CNN OBB}
\title{Sparse R-CNN OBB: Ship Target Detection in SAR Images Based on Oriented Sparse Learnable Proposals}
\name{Kamirul Kamirul \qquad Odysseas Pappas \qquad Alin Achim
\thanks{Kamirul Kamirul is fully funded by Lembaga Pengelola Dana Pendidikan (LPDP), Ministry of Finance of Republic of Indonesia.}
\thanks{
Kamirul Kamirul, Odysseas A. Pappas, and Alin M. Achim are with the
Visual Information Laboratory, University of Bristol, BS1 5DD Bristol,
U.K. (e-mail: kamirul.kamirul@bristol.ac.uk; o.pappas@bristol.ac.uk;
alin.achim@bristol.ac.uk). (Corresponding author: Kamirul Kamirul)
}
}
\address{Visual Information Laboratory\\University of Bristol, UK\\
\{kamirul.kamirul, o.pappas, alin.achim\}@bristol.ac.uk}
\begin{document}
%\ninept
%
\maketitle
\begin{abstract}
We present \ModelName, a novel framework for the detection of oriented objects in SAR images leveraging sparse learnable proposals. 
The \ModelName~has streamlined architecture and ease of training as it utilizes a sparse set of 300 proposals instead of training a proposals generator on hundreds of thousands of anchors.
To the best of our knowledge,~\ModelName~is the first to adopt the concept of sparse learnable proposals for the detection of oriented objects, as well as for the detection of ships in Synthetic Aperture Radar (SAR) images.
The detection head of the baseline model, Sparse R-CNN, is re-designed to enable the model to capture object orientation. 
We train the model on RSDD-SAR dataset and provide a performance comparison to state-of-the-art models.
Experimental results show that~\ModelName~achieves outstanding performance, surpassing most models on both inshore and offshore scenarios. 
The code is available at: \textcolor{blue}{\url{www.github.com/ka-mirul/Sparse-R-CNN-OBB}}. 
\end{abstract}
\section{Introduction}
\label{sec:Introduction}

Synthetic Aperture Radar (SAR) is a powerful active microwave imaging technology that functions reliably under all weather conditions and at any time of day. 
Among its various applications in maritime scenarios, ship detection is one of the most prevalent, applicable for both civilian and military purposes.

%\subsection{Current Research}
The ship detection task involves localizing ships within an image and determining their classes. 
In traditional SAR ship detection systems, the process generally involves three key stages: pre-processing, candidate extraction, and discrimination.
The pre-processing step may involve enhancing the quality of the input image, reducing noise, and conducting land-sea segmentation to avoid extracting non-ocean targets.
In candidate extraction stage, potential ship targets are identified. Methods based on the Constant False Alarm Rate (CFAR) technique are commonly used for this task.
CFAR-based detectors are adaptive threshold systems that estimate sea clutter statistics surrounding a potential target using an assumed background probability density function. 
This estimation enables them to maintain a consistent, acceptable probability of false alarm (PFA) \cite{SP-CFAR_Odysseas}.
Finally, the discrimination stage ensures that only genuine ships are retained for further analysis.
% This step commonly achieved through utilization of artificially designed features, followed by the application of machine learning classifiers. 
Although traditional detectors generally deliver satisfactory performance, their accuracy is prone to variability due to their heavy reliance on hand-crafted parameters assigned at each stage.
Furthermore, the lack of end-to-end design in these approaches increases design complexity, making both training and adaptation to other datasets more challenging.

With the advancement of Convolutional Neural Networks (CNNs), there is a growing trend toward their use in developing Oriented Bounding Box (OBB) detectors for oriented ship detection in SAR images.
Generally, oriented SAR ship detectors draw inspiration from detectors for common oriented objects which can be categorized into three classes, one-stage, two-stage, and anchor-free detectors.
One stage and two-stage detectors are also known as anchor-based approaches due to their dependence on pre-defined anchor boxes. 
The aim of model training is to regress these boxes to accurately align with object boundaries. 
Unlike two-stage detectors that employ a small network to first extract the object candidates (also called proposals), one-stage detectors directly regress the anchor boxes.
One-stage detectors include R-RetinaNet~\cite{R-RetinaNet}, S2ANet~\cite{S2ANet}, and R3Det~\cite{R3Det}, while two-stage detectors include Gliding Vertex~\cite{Gliding_Vertex}, Oriented RCNN~\cite{Oriented_RCNN}, and ReDet~\cite{ReDet}. 
Additionally, CFA~\cite{CFA} and BBAV~\cite{BBAV} are anchor-free detectors that perform predictions directly without relying on predefined anchor boxes.

Overall, anchor-based approaches remain favorable due to their superior performance.
However, the use of dense anchors introduces persistent challenges such as resulting in redundant outputs and causing model performance to be strongly dependent on the initial anchor configurations. 
These challenges have led to the utilisation of fewer anchors, introducing the concept of sparse proposals.

%\subsection{Contributions}

In this work, we introduce~\ModelName, a new family of Region-based Convolutional Neural Networks (R-CNN) utilizing rotated sparse proposals, for the detection of oriented ships in SAR images.
\ModelName~eliminates the need for hundreds of thousands of anchors.
Instead, it utilizes only 300 identically generated learnable proposals.
Our main contributions in this article are twofold. First, we developed~\ModelName, first of its kind to adopt sparse learnable proposals for the detection of oriented objects as well as for the detection of ships in SAR images. 
Second, we fine-tuned the model on oriented SAR ship dataset, RSDD-SAR~\cite{RSDD-SAR}, and provide comparisons to state-of-the-art models.

The remainder of this article is organized as follows, Section \ref{Section:RelatedWork} explains the concept of sparse learnable proposals. 
Section \ref{Section:ProposedModel} provides the implementation details of~\ModelName~. 
Experimental details are provided in Section \ref{Section:ExperimentSetup}, while the experimental results and performance comparison to other models are covered in Section \ref{Section:ResultsandDiscussion}.
Finally, Section \ref{Section:Conclusion} provides concluding remarks.

\section{Related Work}
\label{Section:RelatedWork}
\subsection{Sparse Learnable Proposals}
The concept of utilizing sparse proposal detectors has been explored in prior research~\cite{GCNN, DETR, Sparse-RCNN, Sparse_Anchoring}, with G-CNN~\cite{GCNN} serving as a pioneering approach in this domain. 
Instead of relying on a proposal generator, G-CNN initiates with a multi-scale grid comprising 180 fixed bounding boxes.
These initial boxes are iteratively refined by a regressor that adjusts their position and scale to better align with the objects in the image. At the time of its introduction, G-CNN achieved performance comparable to Fast R-CNN~\cite{Fast-RCNN-original-paper}, which operates with approximately 2,000 bounding boxes. However, G-CNN lags behind the performance of its successor, Faster R-CNN~\cite{Faster-RCNN-original-paper}.

DETR~\cite{DETR} advanced the field by proposing a transformer-based architecture that redefines object detection as a set prediction problem. 
DETR employs a fixed set of 100 learned object queries to simultaneously capture object relationships and global image context, enabling direct parallel predictions. 
Despite its conceptual elegance, DETR relies on dense interactions between object queries and global image features, resulting in slow convergence during training and hindering the development of a fully sparse detection pipeline

Sparse R-CNN~\cite{Sparse-RCNN} is the first framework to introduce a fully sparse proposal paradigm for object detection.
Instead of relying on extensive candidate boxes or dense interactions with global image features, Sparse R-CNN employs a compact set of learnable proposals, known as sparse learnable proposals, comprising both bounding box coordinates and feature embeddings. 
These proposals interact directly with Region of Interest (RoI) features in a one-to-one manner, eliminating the need for global context processing. 
This fully sparse interaction significantly reduces computational complexity while achieving efficient and precise object detection.
The sparse learnable proposals in Sparse R-CNN provide key advantages over traditional dense proposals. They eliminate the need for a region proposal network (RPN) during training and non-maximum suppression (NMS) during inference as densely-overlapping predictions are inherently avoided. 
This results in a simplified architecture comprising only a backbone feature extractor, an interaction module, and classification/regression heads

The Sparse Anchoring Network (SAN) proposed in~\cite{Sparse_Anchoring} represents another effort to incorporate sparse proposals by generating a reduced set of region proposals. However, SAN operates as a region proposal network (RPN) within the R-CNN framework, preventing it from achieving the simplicity and efficiency of the fully sparse design of Sparse R-CNN.

While the architecture of~\ModelName~ will be introduced at a later stage, it is worth noting here that the training regime of~\ModelName~is similar to the second stage of Faster R-CNN~\cite{Faster-RCNN-original-paper} training. 
After feature maps generated by the backbone network, the Region of Interest (RoI) features are extracted through RoIPooling operation on one of the map.
The key difference is that, unlike Faster R-CNN, which directly feeds RoI features to the classification and regression branches,~\ModelName~first interacts them with proposal features through dynamic head. 

\subsection{Dynamic Head}

The dynamic head facilitates the interaction between the proposal features and pooled RoI features, yielding the final object features. 
This interaction is realized through two consecutive 1x1 convolution layers with ReLU activation functions.
The resulting object features are then passed to the regression and classification branches to obtain box offsets and class logits for the proposals. 
The box offsets are then used to update proposal box parameters.

In the original implementation of Sparse R-CNN, a stack of dynamic head was utilized to enhance the model performance. 
In this schema, the updated box parameters and the resulted object features are fed into the subsequent head, forming an iterative refinement process.
Following this strategy, our proposed~\ModelName~employs the same approach by utilizing a stack of six dynamic heads.

\section{\ModelName}
\label{Section:ProposedModel}
This section presents the details of the proposed model along with the strategy used to incorporate the orientation into the model.

\subsection{Backbone}
\ModelName~employs a ResNet-50 backbone with a Feature Pyramid Network (FPN) for multi-scale feature fusion, referred to as ResNet-50-FPN.
As depicted in Fig.~\ref{fig:R-Sparse_R-CNN_Flowchart}, residual blocks of ResNet-50 (\textit{res2}, \textit{res3}, \textit{res4}, \textit{res5}) first output feature maps C2, C3, C4, and C5, which are then passed to the FPN. 
Afterwards, to standardize their channel dimensions (256, 512, 1024, 2048), a 1×1 lateral convolution is applied to align all these maps to 256 channels. 
The aligned maps are further iteratively fused in a top-down manner, with smaller maps upsampled to match the larger ones. Finally, a final 3×3 convolution refines the fused maps, producing outputs P2, P3, P4, and P5.

\subsection{Rotated Sparse Learnable Proposals}
\begin{figure}[h]
\includegraphics[width=0.18\textwidth,center]{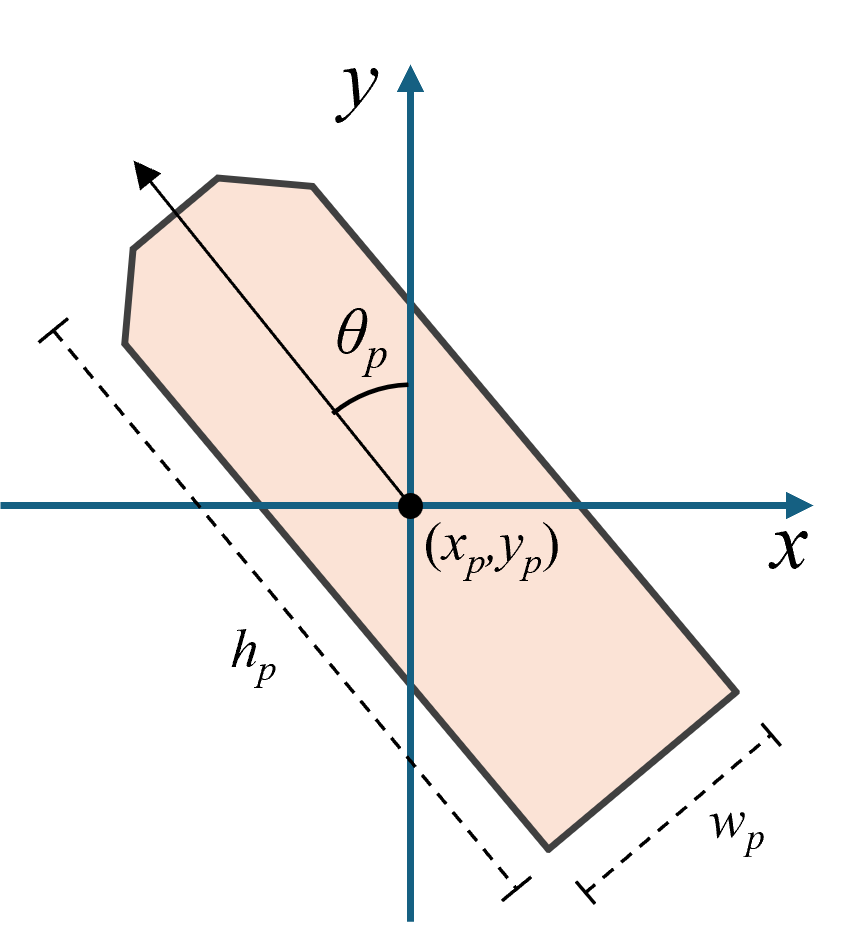}
\caption{\small \colA{Oriented bounding box representation used in the proposed~\ModelName.}}
\label{fig:Ship_Coordinate}
\end{figure}

We exploit the concept of sparse learnable proposals, originally built for detecting non-rotated objects, to further accommodate the detection of oriented objects.
In our pipeline, each proposal contains 256-dimensional learnable proposal features and 5-dimensional box parameters $x_p$, $y_p$, $w_p$, $h_p$, $\theta_p$. 
As depicted in Fig.~\ref{fig:Ship_Coordinate}, the ($x_p$, $y_p$) represents the center point of the proposal while ($w_p$, $h_p$, $\theta_p$) respectively represent the width, height, and orientation of the proposal. 

\begin{figure}[t]
\includegraphics[width=0.45\textwidth,center]{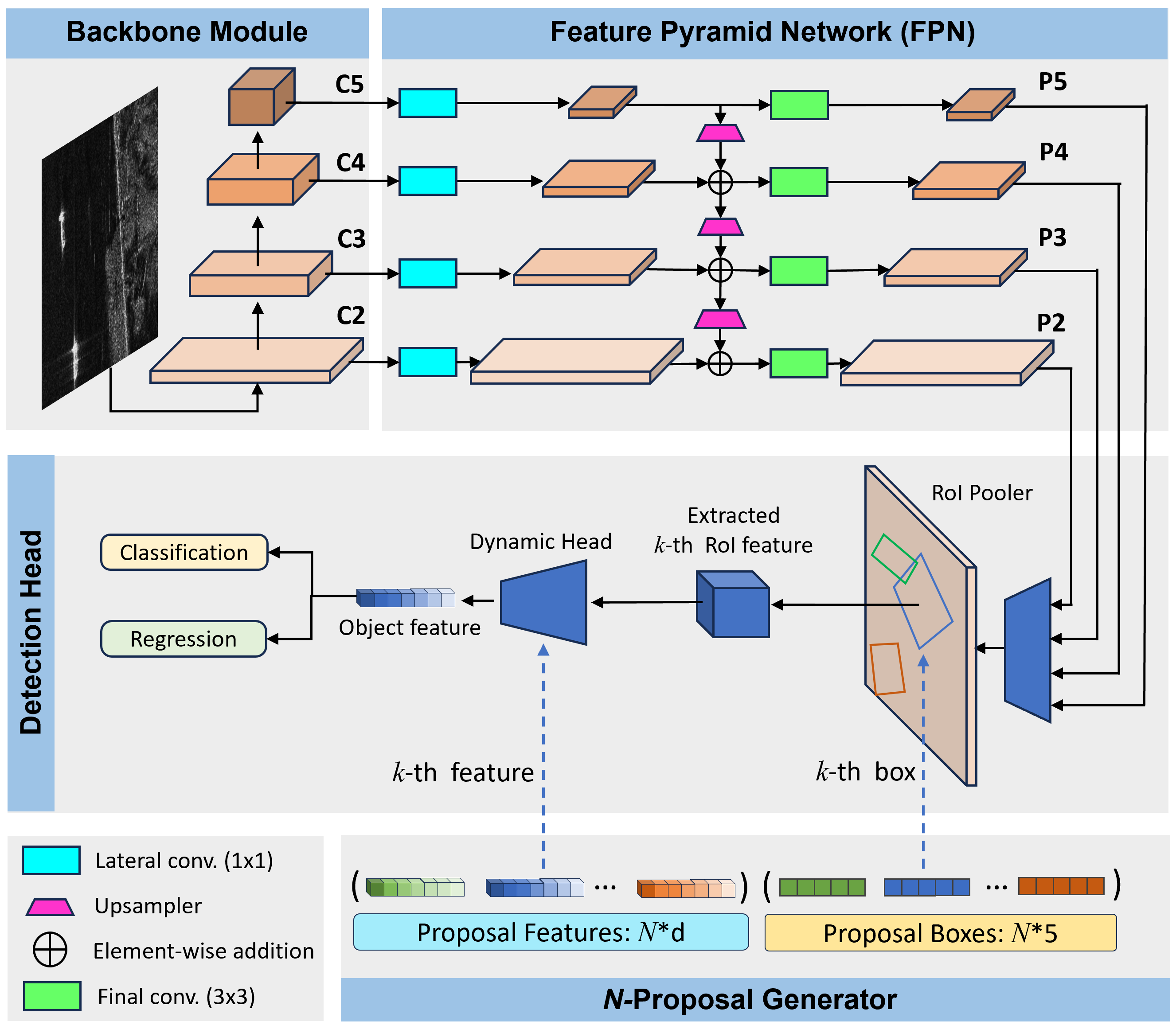}
\caption{\small Flowchart of the proposed \ModelName.}
\label{fig:R-Sparse_R-CNN_Flowchart}
\end{figure}

To accommodate the embedding of the orientation parameter $\theta_p$, we re-designed the baseline model~\cite{Sparse-RCNN} with two significant adjustments.
First, in the RoI pooling stage, instead of using standard RoIAlign~\cite{Mask_RCNN}, we used Rotated RoIAlign (R-RoIAlign)~\cite{Oriented_RCNN} on the feature maps outputted by ResNet-50-FPN. 
This is intended to better capture the feature of the targets enclosed by rotated proposal boxes. 
Secondly, we modified the structure of the regression layers so that they output five parameters instead of four, representing the offset of each rotated box parameter: \(\delta_{x}\), \(\delta_{y}\), \(\delta_{w}\), \(\delta_{h}\), and \(\delta_{\theta}\). The updated box parameters are then obtained by following equation:

\begin{equation}
\begin{gathered}
\hat{x} = x_p + \delta_x \cdot w_p \cdot \cos\theta_p + \delta_y \cdot h_p \cdot \sin\theta_p,\\
\hat{y} = y_p + \delta_x \cdot w_p \cdot \sin\theta_p + \delta_y \cdot h_p \cdot \cos\theta_p,\\
\hat{w} = w_p \cdot e^{\delta_w},\\
\hat{h} = h_p \cdot e^{\delta_h},\\
\hat{\theta} = \theta_p + \delta_\theta,
\end{gathered}
\end{equation} where ($\hat{x}$, $\hat{y}$, $\hat{w}$, $\hat{h}$,  $\hat{\theta}$) represent the updated proposal box parameters. 
The schematic of proposed \ModelName~is provided in Fig.~\ref{fig:R-Sparse_R-CNN_Flowchart}.

As reported in~\cite{Sparse-RCNN}, the initial proposal box parameters ($x_p$, $y_p$, $w_p$, $h_p$, $\theta_p$) have negligible effect on performance. Therefore, we initialize them by placing boxes at the image center, with width and height set to \(\frac{1}{4}\) and \(\frac{1}{2}\) of the image size, and an orientation of \(-\frac{\pi}{4}\).

\subsection{Loss Function}

In ~\ModelName~training, two types of loss calculations are performed: matching loss and training loss. 
The matching loss measures the differences between proposals and the ground truth values. 
% A smaller matching cost indicates that the proposal closely match with the ground truth box. 
The matching cost is formulated as:

\begin{equation}
    \mathcal{L} = \lambda_{\text{cls}} \cdot \mathcal{L}_{\text{cls}} + \lambda_{\text{L1}} \cdot \mathcal{L}_{\text{L1}} + \lambda_{\text{iou}} \cdot \mathcal{L}_{\text{iou}}
\end{equation}
\colA{
In this formulation, the total loss is a weighted sum of the focal loss $\mathcal{L}_{\text{cls}}$, L1 bounding box regression loss $\mathcal{L}_{\text{iou}}$, and IoU loss $\mathcal{L}_{\text{iou}}$, with coefficients $\lambda_{\text{cls}}, \lambda_{\text{L1}}, \lambda_{\text{iou}}$. 
The $\lambda_{\text{cls}}$, $\lambda_{\text{L1}}$, and $\lambda_{\text{IoU}}$ coefficients were set to 2.0, 5.0, and 2.0, respectively, following the baseline Sparse R-CNN settings~\cite{Sparse-RCNN}.
}
The training loss, computed only for matched pairs, follows the same structure as the matching loss and is normalized by the number of objects in the batch.

\section{Experiment Setup}
\label{Section:ExperimentSetup}
\subsection{Dataset}
The~\ModelName~is trained and evaluated on the RSDD-SAR dataset~\cite{RSDD-SAR}, specifically designed for tuning oriented SAR ship detector. 
The dataset contains 7,000 images (512×152 pixels) from TerraSAR-X and Gaofen-3 satellites, with 10,263 annotated ships. 
It covers spatial resolutions from 2 to 20 meters and includes multiple polarization modes (HH, HV, VH, VV, DH, DV). 
RSDD-SAR provides both inshore and offshore test sets for evaluating model performance in different environments.

\subsection{Evaluation}
% read practical : https://learnopencv.com/mean-average-precision-map-object-detection-model-evaluation-metric/

In this work, average precision ($AP$)~\cite{MS_COCO} is utilized as evaluation metric. $AP$ is defined as the area under the \textit{Precision}–\textit{Recall} (\textit{PR}) curve formulated as:

\begin{equation}
AP = \int_{0}^{1} \text{P}(\text{R}) \, d(\text{R}).
\end{equation}
The \textit{Precision} (\textit{P}) and \textit{Recall} (\textit{R}) are defined as follow:

\begin{equation}
Precision = \frac{N_{TP}}{N_{GT}}
\end{equation}

\begin{equation}
Recall =  \frac{N_{TP}}{N_{DET}}.
\end{equation}
Here, $N_{TP}$, $N_{GT}$, and $N_{DET}$ represent the number of true positive predictions, ground truth boxes, and predictions made by the model, respectively. 
Furthermore, Intersection over Union (IoU) is employed to determine true positive detections.
Specifically, we used $AP_{50}$, which requires a minimum IoU score of 0.5 between predicted and ground truth boxes to assign a prediction as a true positive.
Additionally, we employed also $AP_{50}Inshore$ and $AP_{50}Offshore$ to evaluate our model in various test sets.

\subsection{Hyperparameters and Environment}
\label{Hyperparameters_and_Environment}
\ModelName~is set up with weights pre-trained on ImageNet~\cite{ImageNet}. 
The model is then trained using the AdamW optimizer~\cite{Adam_Optimizer} with a base learning rate of \(7.5 \times 10^{-5}\) for 150 epochs, with momentum set to 0.9, weight decay set to \(1 \times 10^{-4}\). 
A warm-up strategy is implemented for the first 1000 iterations, with a learning rate 1\% of the base rate. 
The base learning rate is further reduced by a factor of 10 at the 130\textsuperscript{th} and 140\textsuperscript{th} epochs. 
Training uses a batch size of 8 on two NVIDIA RTX 2080 GPUs with Detectron2~\cite{Detectron2} in PyTorch on Ubuntu 22.04, hosted by the Advanced Computing Research Centre's HPC systems at the University of Bristol.

\section{Results and Discussion}
\label{Section:ResultsandDiscussion}
In this section, we present the experimental results and compare the performance of \ModelName~ against state-of-the-art (SOTA) models.

\subsection{Effect of Number of Proposals}

We conducted an experiment by varying the number of proposals to asses their effects on accuracy, model size, training time, and inference speed.

\begin{table}[htbp]
\centering
\caption{Experiment on Number of Proposals}
\label{table:Exp_Number_of_Proposal}
\renewcommand{\arraystretch}{1.2} % Adjust the value to increase the spacing
\setlength{\tabcolsep}{2.0pt}
\begin{tabular}{>{\centering\arraybackslash}m{1.5cm}>{\centering\arraybackslash}m{1.1cm}>{\centering\arraybackslash}m{2.0cm}>{\centering\arraybackslash}m{1.9cm}>{\centering\arraybackslash}m{1.3cm}}
\hline
\textbf{Proposals}& 
\textbf{AP\textsubscript{50}(\%)}& 
\textbf{Model size (M)}& 
\textbf{Training Duration (h)}&
\textbf{FPS} \\\hline

100 & 91.02 & 106.13 & 11.63 & 14 \\ 
200 & 91.32 & 106.15 & 13.34 & 12\\ 
300 & 91.78 & 106.18 & 14.60 & 11\\ \hline
\end{tabular}
\raggedright
\end{table}

% The observed performance gains with more proposals suggest that \ModelName~can be tailored to scenarios specific tasks depending on the desired accuracy level and training time constraints. 
According to Table \ref{table:Exp_Number_of_Proposal}, increasing the number of proposals marginally enhances performance, but at the cost of longer training times and reduced inference speed. 
The results also demonstrate that using more proposals does not significantly increase model size.
This is because the increase in proposals only adds parameters for storing the 5-dimensional proposal boxes and 256-dimensional proposal features, which are minimal compared to the overall model size.

Considering the above performance, the 300-proposal configuration is set as the default for the model, employed for the rest of the experiments.
However, it should be noted that the number of proposals limits the maximum detectable objects and should be adjusted based on the application. For instance, using 100 proposals would be inappropriate for detecting hundreds of objects in a single image.

\subsection{Comparison to SOTA Models}
In this section, we compare the performance of~\ModelName~against state-of-the-art algorithms.
As shown in Table \ref{table:performance_comparison_RSDD-SAR},~\ModelName~outperforms other methods across all metrics, except for $AP_{50}Inshore$.
In mixed-scene test sets, reflected by {\bf $AP_{50}$}, our method demonstrates superior performance, surpassing all state-of-the-art algorithms by a significant margin ranging from 2.51\% to 25.16\%, with CFA~\cite{CFA} being the closest competitor.
Compared to other two-stage detectors, our model demonstrates a superior performance with an improvement of 2.98\% over Oriented R-CNN \cite{Oriented_RCNN}, the top-performing algorithm in this category. 
This performance comparison shows that~\ModelName~is robust for detecting ships in typical operational ocean environments, where both inshore and offshore backgrounds may coexist.

\begin{figure*}[!htbp]
\includegraphics[width=0.86\textwidth,center]{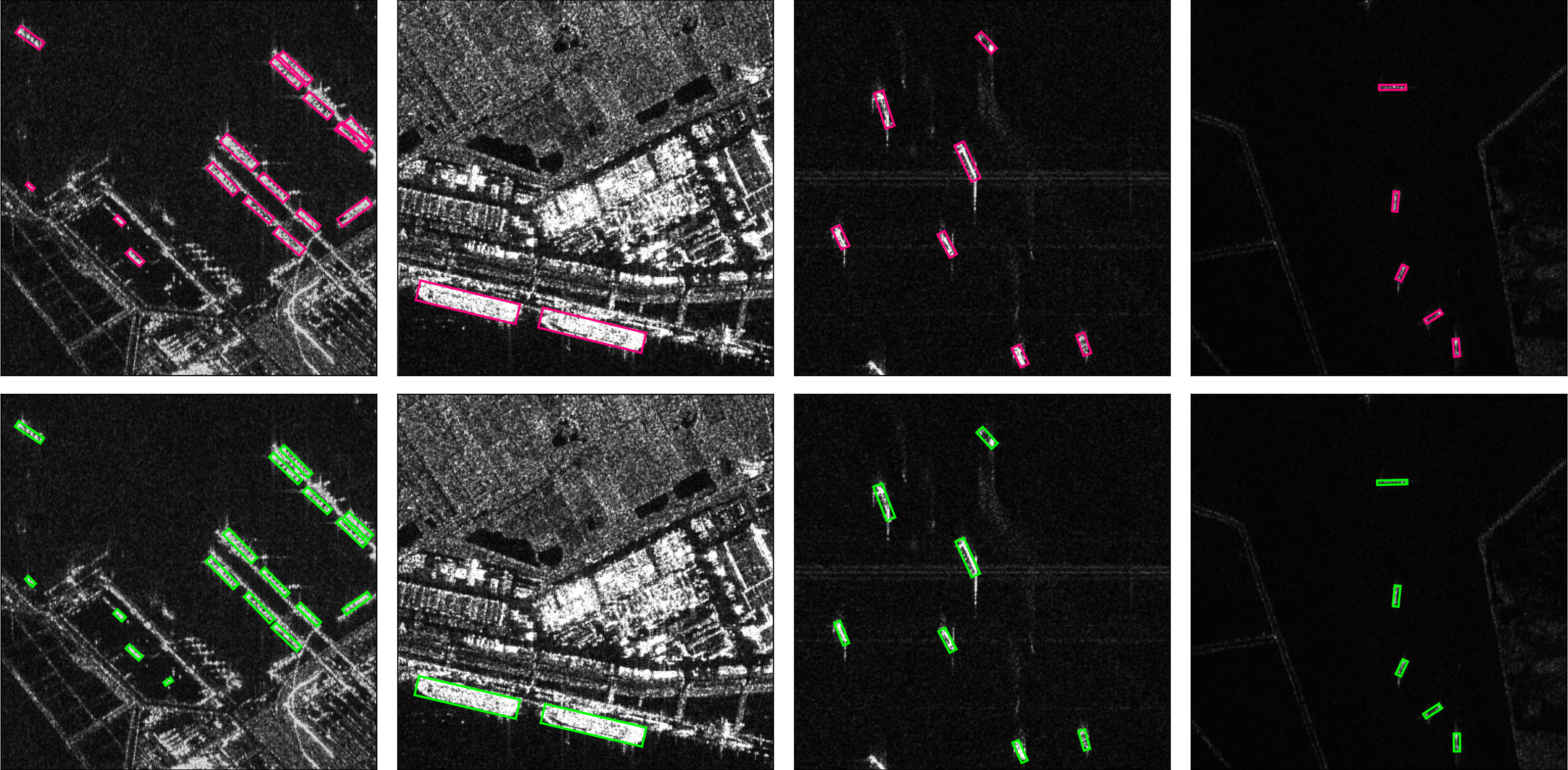}
\caption{\small
Detection results across different environments, with predictions (magenta) and ground truth (green). 
The first two left columns show results in inshore regions, while the remaining columns correspond to offshore regions.}
\label{fig:SampleDetection}
\end{figure*}

\begin{table}[ht]
\centering
\caption{Performance Comparison on RSDD-SAR Dataset}
\label{table:performance_comparison_RSDD-SAR}

\renewcommand{\arraystretch}{1.2} % Adjust the value to increase the spacing
\setlength{\tabcolsep}{3.0pt}
\begin{tabular}{>
{\centering\arraybackslash}m{1.6cm}>
{\centering\arraybackslash}m{2.4cm}>
{\centering\arraybackslash}m{0.7cm}>
{\centering\arraybackslash}m{1.3cm}>
{\centering\arraybackslash}m{1.5cm}}
\hline
\textbf{Type}& 
\textbf{Model}& 
\textbf{AP\textsubscript{50} (\%)}& 
\textbf{AP\textsubscript{50} Inshore (\%)}& 
\textbf{AP\textsubscript{50} Offshore (\%)}\\\hline

\multirow{4}{*}{Two-stage} & R-Faster R-CNN & 83.29 & 48.78 & 90.93 \\ 
 & RoI Transformer & 88.39 & 60.83 & \textbf{94.35*} \\ 
 & Gliding Vertex & 85.55 & 55.93 & 91.65 \\ 
 & Oriented R-CNN & 88.84 & 65.92 & 90.21 \\ \hline
 
\multirow{3}{*}{One-stage} & R-RetinaNet & 66.66 & 33.20 & 74.06 \\ 
 & S2ANet & 87.91 & 63.27 & 93.14 \\ 
 & R3Det & 80.87 & 56.87 & 90.16 \\ \hline

\multirow{3}{*}{Anchor-free} & Polar Encoding & 87.31 & 59.69 & 90.12 \\ 
 & R-FCOS & 85.48 & 50.02 & 93.09 \\ 
 & CFA & \textbf{89.31*} & \textbf{66.40$^\dagger$} & 90.47 \\\hline

\multirow{4}{*}{} & \ModelName & \textbf{91.82$^\dagger$} & \textbf{66.27*} & \textbf{96.26$^\dagger$} \\ \hline

\end{tabular}
\raggedright
\footnotesize{
\textbf{$^\dagger$} Highest value, \textbf{*} Second highest value.}
\\
The performance of other models were taken from~\cite{RSDD-SAR}.
\end{table}

\colB{
In inshore scenes,~\ModelName~achieves performance comparable to CFA, with a marginal gap of less than 0.02\%. 
While CFA is tailored for densely packed, arbitrarily oriented objects using convex-hull-based adaptation,~\ModelName~is specifically designed for SAR ship detection across both inshore and offshore settings. 
These results suggest that incorporating inshore-specific refinements could further enhance performance. 
Nonetheless,~\ModelName~remains highly competitive, exhibiting strong robustness and lower false detection rates in complex environments.}
This result indicates that~\ModelName~can better detect ships in complex environments, whereas other methods struggle with higher false detection rates.

Lastly, for offshore scenes,~\ModelName~demonstrates exceptional performance, achieving more than 96\% accuracy and surpassing other models by at least a 1.91\% margin.
This results signifies that~\ModelName~is more robust against noisy SAR ocean images, which may also be interfered by sea waves and ship-generated wakes.

Finally, Fig.~\ref{fig:SampleDetection} presents sample prediction results from \ModelName~on both datasets across different background scenarios.
These results visually confirm that the proposed model accurately detects most ship targets in accordance with their ground truth, demonstrating its robust performance in varying scenarios.
We anticipate that our \ModelName~presents a versatile solution for detection of oriented ships, delivering enhanced accuracy while maintaining ease of design and training.

\section{Conclusion}
\label{Section:Conclusion}
In this article, we proposed \ModelName, a new pipeline for oriented ship detection in SAR imageries based on sparse learnable proposals. 
The adoption of sparse proposals concept ensures a simple design for our model.
Furthermore, the incorporation of rich instance features from learnable proposals improves the model performance.
~\ModelName~achieves superior performance, surpassing all state-of-the-art benchmarks on the RSDD-SAR dataset by at least 2.51\% in $AP_{50}$.

\bibliographystyle{IEEEbib}
\bibliography{strings,refs}

\begin{thebibliography}{10}

\bibitem{SP-CFAR_Odysseas}
Odysseas Pappas, Alin Achim, and David Bull,
\newblock ``Superpixel-level {CFAR} detectors for ship detection in sar imagery,''
\newblock {\em IEEE Geosci. and Remote Sens. Lett.}, vol. 15, no. 9, pp. 1397--1401, 2018.

\bibitem{R-RetinaNet}
Rong Yang, Gui Wang, Zhenru Pan, Hongliang Lu, Heng Zhang, and Xiaoxue Jia,
\newblock ``A novel false alarm suppression method for {CNN}-based {SAR} ship detector,''
\newblock {\em IEEE Geosci. Remote Sens. Lett.}, vol. 18, no. 8, pp. 1401--1405, 2021.

\bibitem{S2ANet}
Jiaming Han, Jian Ding, Jie Li, and Gui-Song Xia,
\newblock ``Align deep features for oriented object detection,''
\newblock {\em IEEE Trans. Geosci. Remote Sens.}, vol. 60, pp. 1--11, 2022.

\bibitem{R3Det}
Xue Yang, Junchi Yan, Ziming Feng, and Tao He,
\newblock ``{R3Det}: {R}efined single-stage detector with feature refinement for rotating object,''
\newblock in {\em Proc. AAAI Conf. Artificial Intel.}, May 2021, vol.~35, pp. 3163--3171.

\bibitem{Gliding_Vertex}
Yongchao Xu, Mingtao Fu, Qimeng Wang, Yukang Wang, Kai Chen, Gui-Song Xia, and Xiang Bai,
\newblock ``Gliding vertex on the horizontal bounding box for multi-oriented object detection,''
\newblock {\em IEEE Trans. Pattern Anal. Mach. Intell.}, vol. 43, no. 4, pp. 1452--1459, 2021.

\bibitem{Oriented_RCNN}
Xingxing Xie, Gong Cheng, Jiabao Wang, Xiwen Yao, and Junwei Han,
\newblock ``{Oriented R-CNN} for object detection,''
\newblock in {\em 2021 IEEE/CVF Int. Conf. Comp. Vis. (ICCV)}, 2021, pp. 3500--3509.

\bibitem{ReDet}
Jiaming Han, Jian Ding, Nan Xue, and Gui-Song Xia,
\newblock ``{ReDet}: {A} rotation-equivariant detector for aerial object detection,''
\newblock in {\em 2021 IEEE/CVF Conf. Comp. Vis. Pattern Recogn. (CVPR)}, 2021, pp. 2785--2794.

\bibitem{CFA}
Zonghao Guo, Chang Liu, Xiaosong Zhang, Jianbin Jiao, Xiangyang Ji, and Qixiang Ye,
\newblock ``{Beyond Bounding-Box}: {C}onvex-hull feature adaptation for oriented and densely packed object detection,''
\newblock in {\em 2021 IEEE/CVF Conf. Comp. Vis. and Pattern Recogn. (CVPR)}, 2021, pp. 8788--8797.

\bibitem{BBAV}
Jingru Yi, Pengxiang Wu, Bo~Liu, Qiaoying Huang, Hui Qu, and Dimitris Metaxas,
\newblock ``Oriented object detection in aerial images with box boundary-aware vectors,''
\newblock in {\em 2021 IEEE Winter Conf. Appl. Comp. Vis. (WACV)}, 2021, pp. 2149--2158.

\bibitem{RSDD-SAR}
Xu~Congan, Su~Hang, Li~Jianwei, Liu Yu, Yao Libo, Gao Long, Yan Wenju, and Wang Taoyang,
\newblock ``{RSDD-SAR}: {R}otated ship detection dataset in {SAR} images,''
\newblock {\em Journal of Radars}, vol. 11, no. 4, 2022.

\bibitem{GCNN}
Mahyar Najibi, Mohammad Rastegari, and Larry~S. Davis,
\newblock ``{G-CNN}: {A}n iterative grid based object detector,''
\newblock in {\em 2016 IEEE Conf. Comput. Vis. Pattern Recognit. (CVPR)}, 2016, pp. 2369--2377.

\bibitem{DETR}
Nicolas Carion, Francisco Massa, Gabriel Synnaeve, Nicolas Usunier, Alexander Kirillov, and Sergey Zagoruyko,
\newblock ``End-to-end object detection with transformers,''
\newblock in {\em Computer Vision -- ECCV 2020}. 2020, p. 213–229, Springer-Verlag.

\bibitem{Sparse-RCNN}
Peize Sun, Rufeng Zhang, Yi~Jiang, Tao Kong, Chenfeng Xu, Wei Zhan, Masayoshi Tomizuka, Lei Li, Zehuan Yuan, Changhu Wang, and Ping Luo,
\newblock ``{Sparse R-CNN}: {E}nd-to-end object detection with learnable proposals,''
\newblock in {\em 2021 IEEE/CVF Conf. Comp. Vis. Pattern Recogn. (CVPR)}, 2021, pp. 14449--14458.

\bibitem{Sparse_Anchoring}
Yongtao Yu, Jun Wang, Hao Qiang, Mingxin Jiang, E~Tang, Changhui Yu, Yongjun Zhang, and Jonathan Li,
\newblock ``Sparse anchoring guided high-resolution capsule network for geospatial object detection from remote sensing imagery,''
\newblock {\em Int. J. Appl. Earth Observ. Geoinf.}, vol. 104, pp. 102548, 2021.

\bibitem{Fast-RCNN-original-paper}
Ross~B. Girshick,
\newblock ``Fast {R-CNN},''
\newblock {\em CoRR}, vol. abs/1504.08083, 2015.

\bibitem{Faster-RCNN-original-paper}
Shaoqing Ren, Kaiming He, Ross~B. Girshick, and Jian Sun,
\newblock ``Faster {R-CNN:} towards real-time object detection with region proposal networks,''
\newblock {\em CoRR}, vol. abs/1506.01497, 2015.

\bibitem{Mask_RCNN}
Kaiming He, Georgia Gkioxari, Piotr Dollár, and Ross Girshick,
\newblock ``{Mask R-CNN},''
\newblock in {\em 2017 IEEE Int. Conf. Comp. Vis. (ICCV)}, 2017, pp. 2980--2988.

\bibitem{MS_COCO}
Tsung-Yi Lin, Michael Maire, Serge Belongie, James Hays, Pietro Perona, Deva Ramanan, Piotr Doll{\'a}r, and C.~Lawrence Zitnick,
\newblock ``Microsoft {COCO}: {C}ommon {O}bjects in {C}ontext,''
\newblock in {\em Computer Vision -- ECCV 2014}, 2014, pp. 740--755.

\bibitem{ImageNet}
Jia Deng, Wei Dong, Richard Socher, Li-Jia Li, Kai Li, and Li~Fei-Fei,
\newblock ``{ImageNet}: {A} large-scale hierarchical image database,''
\newblock in {\em 2009 IEEE Conf. Comp. Vis. Pattern Recogn. (CVPR)}, 2009, pp. 248--255.

\bibitem{Adam_Optimizer}
Diederik Kingma and Jimmy Ba,
\newblock ``Adam: A method for stochastic optimization,''
\newblock in {\em Int. Conf. on Learning Representations (ICLR)}, San Diega, CA, USA, 2015.

\bibitem{Detectron2}
Yuxin Wu, Alexander Kirillov, Francisco Massa, Wan-Yen Lo, and Ross Girshick,
\newblock ``Detectron2,'' \url{https://github.com/facebookresearch/detectron2}, 2019.

\end{thebibliography}

\end{document}